\documentclass[12pt]{article}

\usepackage{graphicx}
\usepackage{multirow}
\usepackage{amsmath,amssymb,amsfonts}
\usepackage{amsthm}
\usepackage{mathrsfs}
\usepackage[title]{appendix}
\usepackage{xcolor}
\usepackage{textcomp}
\usepackage{manyfoot}
\usepackage{booktabs}
\usepackage{algorithm}
\usepackage{algorithmicx}
\usepackage{algpseudocode}
\usepackage{listings}
\usepackage{makecell}
\usepackage[switch]{lineno}
\usepackage[utf8]{inputenc}
\usepackage{array}
\usepackage{siunitx}
\usepackage{ulem}
\usepackage{hyperref}
\usepackage{cite}
\usepackage{threeparttable}
\usepackage{fullpage}
\usepackage{bbm}

\begin{document}


\title{GPAIR: Gaussian-Kernel-Based Ultrafast 3D Photoacoustic Iterative Reconstruction}

\author{
Yibing Wang\textsuperscript{1} \and
Shuang Li\textsuperscript{1} \and
Tingting Huang\textsuperscript{1} \and
Yu Zhang\textsuperscript{1} \and
Chulhong Kim\textsuperscript{2-6} \and
Seongwook Choi\textsuperscript{2-6} \and
Changhui Li\textsuperscript{1,7}\thanks{Corresponding author: \texttt{chli@pku.edu.cn}}
}

\date{\small
\textsuperscript{1}Department of Biomedical Engineering, College of Future Technology, Peking University, Beijing, China\\
\textsuperscript{2}Department of Electrical Engineering, Pohang University of Science and Technology, Pohang, Republic of Korea\\
\textsuperscript{3}Department of Convergence IT Engineering, Pohang University of Science and Technology, Pohang, Republic of Korea\\
\textsuperscript{4}Department of Mechanical Engineering, Pohang University of Science and Technology, Pohang, Republic of Korea\\
\textsuperscript{5}Department of Medical Science and Engineering, Pohang University of Science and Technology, Pohang, Republic of Korea\\
\textsuperscript{6}Medical Device Innovation Center, Pohang University of Science and Technology, Pohang, Republic of Korea\\
\textsuperscript{7}National Biomedical Imaging Center, Peking University, Beijing, China
}

\maketitle

\abstract{Although the iterative reconstruction (IR) algorithm can substantially correct reconstruction artifacts in photoacoustic (PA) computed tomography (PACT), it suffers from long reconstruction times, especially for large-scale three-dimensional (3D) imaging in which IR takes hundreds of seconds to hours. The computing burden severely limits the practical applicability of IR algorithms. In this work, we proposed an ultrafast IR method for 3D PACT, called Gaussian-kernel-based Ultrafast 3D Photoacoustic Iterative Reconstruction (GPAIR), which achieves orders-of-magnitude acceleration in computing. GPAIR transforms traditional spatial grids with continuous isotropic Gaussian kernels. By deriving analytical closed-form expression for pressure waves and implementing powerful GPU-accelerated differentiable Triton operators, GPAIR demonstrates extraordinary ultrafast sub-second reconstruction speed for 3D targets containing 8.4 million voxels in animal experiments. This revolutionary ultrafast image reconstruction enables near-real-time large-scale 3D PA reconstruction, significantly advancing 3D PACT toward clinical applications.}

\newpage

\section{Introduction}

Three-dimensional photoacoustic computed tomography (3D PACT), combining optical contrast with ultrasonic resolution and significant imaging depth, has established itself as a powerful modality for volumetric biological imaging~\cite{wang2012photoacoustic,assi2023review, dean2017advanced, lin2022emerging, park2025clinical, yang2025multiplane}. Modern 3D PACT systems, employing arrays consisting of hundreds to thousands of elements, have shown immense potential in wide-ranging pre-clinical and clinical studies~\cite{sun2024real, li2024photoacoustic, matsumoto2018visualising, han2021three, lin2021high, kim2023wide, kim2025non, choi2026photoacoustic}. However, constrained by system manufacturing costs and physical restrictions on limited detection aperture during in vivo imaging, 3D PACT systems generally have sparse sensors and limited-view detection geometries. These limitations inevitably compromise signal acquisition and cause severe artifacts when using standard back-projection-based reconstruction algorithms like universal back-projection (UBP)~\cite{xu2005universal}.

To address these hardware constraints, two primary strategies are being explored: mechanical scanning and advanced reconstruction algorithms. While mechanical scanning can effectively improve spatial sampling density~\cite{na2022massively, li2024photoacoustic}, it severely compromises the imaging speed. At the same time, various advanced reconstruction algorithms have gained much progresses. Physics-based iterative reconstruction (IR), which numerically solves wave propagation equation (e.g., k-Wave), provides high fidelity but suffers from prohibitive memory and time costs~\cite{rosenthal2010fast, dean2012accurate, wang2014discrete, zhu2023mitigating, treeby2010k}. To accelerate this, model-based semi-analytical approaches, like model-based point-detector algorithm (MB-PD)~\cite{ding2017efficient, ding2020model}, were developed to minimize reconstruction time. However, these methods typically rely on a ``discrete-to-discrete'' voxel paradigm that introduces discretization errors. Concurrently, data-driven deep learning (DL) approaches have emerged, offering rapid inference speeds~\cite{zheng2023deep, he2025pacformer, jeong2025hybrid, wang2025msd}. Yet, DL-based algorithms often lack the rigorous physical interpretability required for medical diagnosis and have risk of hallucination artifacts when the imaging domain shifts from the training distribution.

Parallel advancements in computer vision have shifted towards continuous scene representations to overcome resolution limits, exemplified by differentiable rendering, neural radiance fields (NeRF), and 3D Gaussian splatting (3DGS)~\cite{mildenhall2021nerf, barron2021mip, weiss2021differentiable, chen2022tensorf, kerbl20233d}. Inspired by these breakthroughs, recent PACT methods like SlingBAG have explored point-based representations to enhance efficiency~\cite{li2025slingbag}. Nevertheless, current state-of-the-art methods still require hundreds of seconds to complete a reconstruction for complex source structures, remaining far from the demands of real-time applications.

In this study, we propose Gaussian-kernel-based Ultrafast 3D Photoacoustic Iterative Reconstruction (GPAIR). By representing the initial pressure field as a superposition of continuous isotropic Gaussian kernels, we derive a closed-form analytical expression for the forward and adjoint operators. This formulation eliminates errors associated with numerical time-stepping or voxel interpolation. We implement these operators using optimized GPU-native Triton kernels~\cite{tillet2019triton} incorporated with an adaptive supersampling alignment (ASSA) strategy to maximize computational throughput. Furthermore, to ensure robustness, GPAIR employs a nonnegative parameterization constraint (NPC) alongside vessel continuity regularization (VCR). Validated on large-scale simulation and in vivo datasets, GPAIR achieves up to 872$\times$ speedup over existing methods while delivering superior reconstruction fidelity.

\newpage

\section{Results}

\subsection{Principle of Gaussian-kernel-based continuous reconstruction}

\begin{figure}[htbp]
\centering
\includegraphics[width=0.95\textwidth]{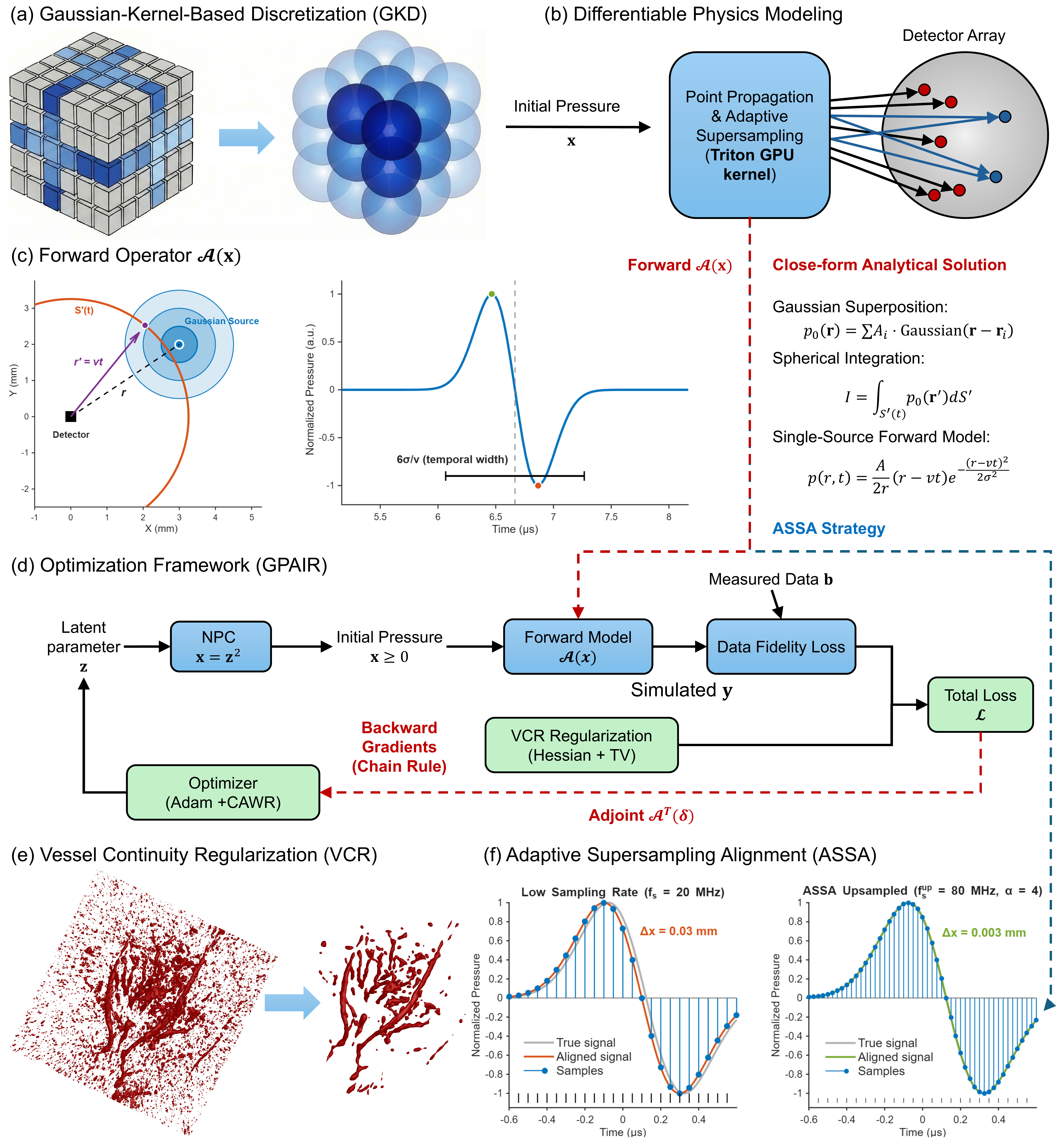}
\caption{Schematic principle of GPAIR. (a) Gaussian-Kernel-Based Discretization (GKD): transforming discrete voxels into continuous Gaussian spheres. (b) Differentiable Physics Modeling: using Triton GPU kernels for efficient point propagation. (c) Forward Operator $\mathcal{A}(\mathbf{x})$: showing the analytical Gaussian waveform and temporal width. (d) Optimization Framework (GPAIR): the iterative loop with NPC ($\mathbf{x}=\mathbf{z}^2$), forward model, VCR regularization, and Adam optimizer. (e) Vessel Continuity Regularization (VCR): enhancing vessel connectivity using Hessian and TV priors. (f) Adaptive Supersampling Alignment (ASSA): correcting discretization misalignment by upsampling ($f_s=20$ MHz vs. $f_s^{up}=80$ MHz).}
\label{fig:methods}
\end{figure}

GPAIR establishes a fully differentiable reconstruction framework based on continuous physics representation. As illustrated in Fig.~\ref{fig:methods}, GPAIR integrates three coupled components: Gaussian-Kernel-Based Discretization (GKD), differentiable physics modeling with Adaptive Supersampling Alignment (ASSA), and a constrained optimization loop with Vessel Continuity Regularization (VCR).

First, regarding the source representation (Fig.~\ref{fig:methods}(a)), GPAIR employs the GKD strategy instead of the discrete voxel model. We model the initial pressure field $p_0(\mathbf{r}) = \sum_i A_i \cdot \text{Gaussian}(\mathbf{r} - \mathbf{r}_i)$ as a linear superposition of continuous isotropic Gaussian basis functions: ${p_0}_i(\mathbf{r}) = A_i \cdot \text{Gaussian}(\mathbf{r} - \mathbf{r}_i)$. This continuous formulation confers infinite spatial differentiability to the source field, fundamentally eliminating high-frequency artifacts introduced by voxel non-smooth boundaries.

Second, for forward wave propagation physics modeling (Fig.~\ref{fig:methods}(b)--(c)), we construct a fully differentiable forward operator $\mathcal{A}(\mathbf{x})$. Leveraging the mathematical properties of Gaussian functions, for acoustically homogeneous and non-viscous medium, a closed-form analytical expression for the generated PA wave from a single Gaussian source can be derived. As shown in Fig.~\ref{fig:methods}(c), the generated PA signal $p(r,t)$ is expressed explicitly as a function of time $t$ and the detector-to-source distance $r$, characterized by an N-shaped-like profile. This analytical approach bypasses the computationally expensive time-stepping. To address the misalignment between the continuous analytical waveform and discrete digital sampling, we introduce the ASSA strategy (Fig.~\ref{fig:methods}(f)), which dynamically calculates an upsampling factor $\alpha$ to perform point propagation on a refined temporal grid, thereby suppressing discretization errors to negligible levels. This entire forward process is implemented via highly parallelized Triton GPU kernels to maximize computational throughput.

Finally, the reconstruction is formulated as a gradient-based iterative optimization loop (Fig.~\ref{fig:methods}(d)). To ensure physical rationality, we employ a Nonnegative Parameterization Constraint (NPC), reparameterizing the pressure $\mathbf{x}$ as $\mathbf{x}=\mathbf{z}^2$ to strictly enforce non-negativity. During iteration, the optimizer updates the latent variable $\mathbf{z}$ by backpropagating gradients (Adjoint $\mathcal{A}^T(\mathbf{\delta})$) derived from the total loss $\mathcal{L}$, which combines data fidelity with regularization. To tackle the ill-posedness of sparse-view imaging, we incorporate the VCR strategy (Fig.~\ref{fig:methods}(e)), which synergizes Hessian-based second-order curvature priors with Total Variation (TV) first-order sparsity. VCR effectively suppresses background noise while preserving the topological connectivity of vascular structures.

In summary, by applying continuous physics modeling with high-performance computing, GPAIR achieves a paradigm shift from ``discrete-to-discrete'' to ``continuous-to-discrete'' mapping, establishing a rigorous theoretical foundation for high-fidelity, ultrafast 3D photoacoustic imaging.

\subsection{Validation on synthetic vascular data}

\begin{figure}[htbp]
\centering
\includegraphics[width=0.85\textwidth]{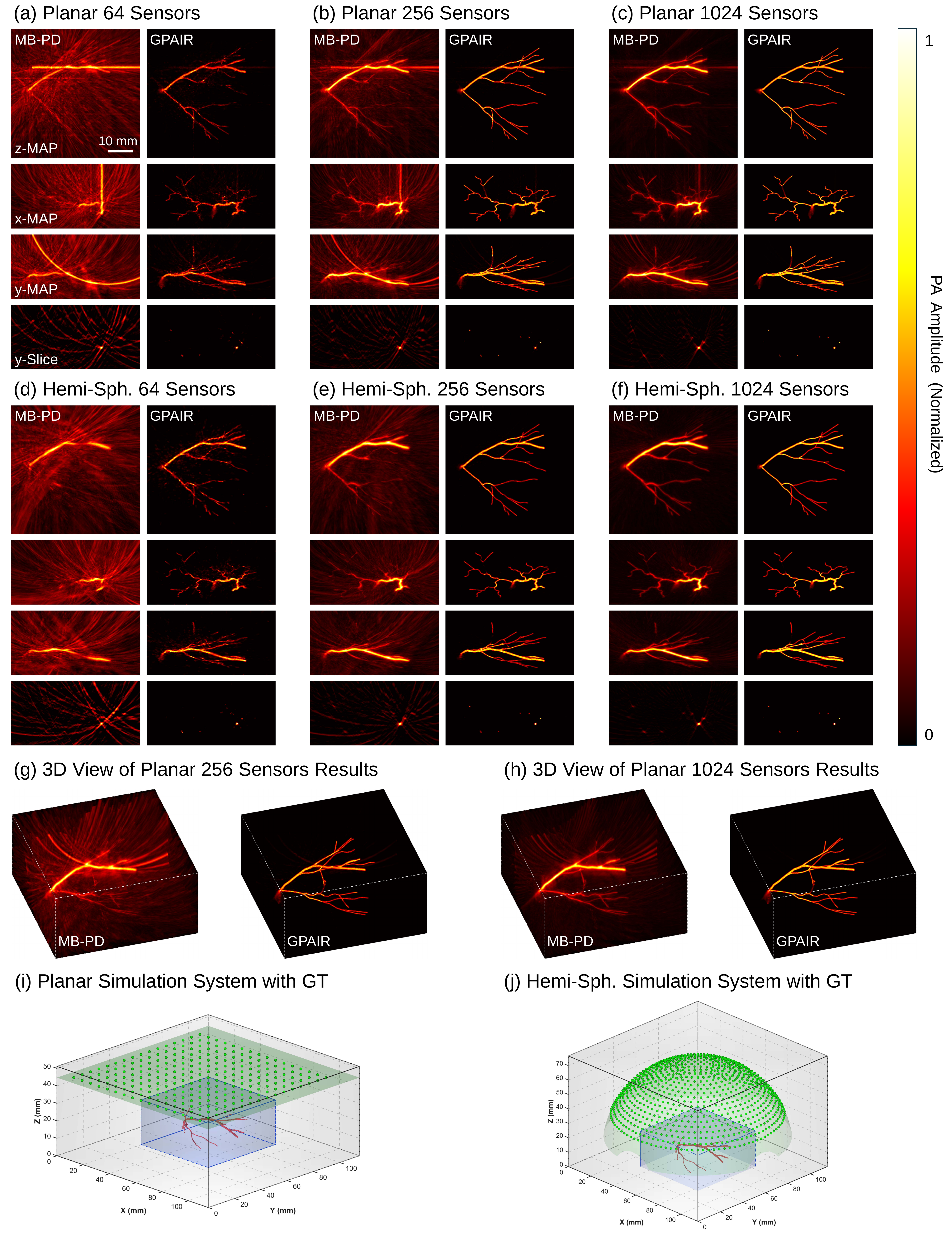}
\caption{Validation using simulation study under two system configurations. (a)--(c) Planar array results with 64, 256, and 1024 sensors, respectively. (d)--(f) Hemispherical array results with 64, 256, and 1024 sensors, respectively. Rows: z-MAP, x-MAP, y-MAP, y-Slice; Columns: MB-PD, GPAIR. (g), (h) 3D volumetric rendering of MB-PD and GPAIR results with planar 256 and 1024 sensors. (i), (j) Planar and hemi-spherical array simulation geometry with GT. Scale bar: 10~mm.}
\label{fig:simulation}
\end{figure}

To systematically evaluate the computational efficiency and reconstruction fidelity of GPAIR, we conducted simulation studies using a synthetic 3D complex vascular network ($512 \times 512 \times 256$ voxels) as the ground truth (GT)~\cite{galarreta2013three}, and its PA pressure wave are generated by the k-Wave toolbox. As illustrated in Fig.~\ref{fig:simulation}, we investigated two representative array geometries: a planar array ($102.4\times102.4$~mm) and a hemispherical array ($R = 60$~mm), each with three detector densities (64, 256, and 1024 elements, respectively). We compared results of maximum amplitude projections (MAP) from three views and one single slice, which are reconstructed by an NVIDIA RTX 5090 GPU (the same of all the following studies).

Fig.~\ref{fig:simulation} presents the comparative reconstruction results between previously reported fast semi-analytical IR method, MB-PD~\cite{ding2017efficient,ding2020model}, and GPAIR. We first analyzed the performance under the planar array geometry (Fig.~\ref{fig:simulation}(a)--(c)). In the extremely sparse case with 64 sensors (Fig.~\ref{fig:simulation}(a)), MB-PD reconstructions were degraded by significant background noise and aliasing artifacts, resulting in discontinuous vessel structures. In contrast, GPAIR maintained high structural integrity, effectively suppressing artifacts and clearly resolving the main vascular trunks. The y-Slice views further demonstrate that GPAIR recovered sharp vascular cross-sections with a clean background. As the detector density increased to 256 elements (Fig.~\ref{fig:simulation}(b)), GPAIR continued to outperform MB-PD, reconstructing a more continuous and complete vascular network from main trunks to fine branches.

Notably, under the 1024-element planar configuration (Fig.~\ref{fig:simulation}(c)), GPAIR uniquely recover vessel along the depth direction (perpendicular to the array plane). These vertical vessels are typically hard to recover for planar geometries due to the limited-view issue. By exploiting advanced spatial continuity constraints, GPAIR successfully inferred these structures. This advantage is visually confirmed by the 3D volumetric renderings in Fig.~\ref{fig:simulation}(g) and (h). While MB-PD results (Fig.~\ref{fig:simulation}(g)) exhibit amplitude loss and missing vertical segments, GPAIR (Fig.~\ref{fig:simulation}(h)) restores the complete vascular topology, matching the ground truth closely.

For the hemispherical array geometry (Fig.~\ref{fig:simulation}(d)--(f)), GPAIR similarly exhibited superior robustness. Even with only 64 sensors (Fig.~\ref{fig:simulation}(d)), GPAIR achieved high-fidelity reconstruction with negligible artifacts, whereas MB-PD suffered from streak noise. Across all sensor densities, GPAIR consistently provided significantly higher contrast and much better vessel connectivity than MB-PD. More detailed multi-view comparisons, including the traditional UBP reconstruction results, are provided in Supplementary Figures 1--3 and Supplementary Note 1, with corresponding 3D visualization videos available in Supplementary Movies 1--3.

Quantitative metrics corroborates these visual findings (Table~\ref{tab:simulation_metrics}). GPAIR consistently achieved the highest PSNR and SSIM across all configurations. Even under extreme 64-element sparse sampling, GPAIR attained a PSNR of over 36~dB, significantly outperforming MB-PD ($\approx 26$~dB) and UBP ($\approx 24$~dB). As shown in Table~\ref{tab:simulation_metrics}, the SSIM of GPAIR remained above 0.99 for all sensor densities, while conventional methods degraded significantly as the number of sensors decreased. Besides, the GPAIR demonstrates at least of two orders' speedup in computing time, making it highly suitable for large-scale 3D PACT applications.

\begin{table}[htbp]
\centering
\small
\setlength{\tabcolsep}{3pt}
\begin{threeparttable}
\caption{Quantitative Metric and Computational Performance Comparison in Simulation Experiments}
\begin{tabular}{ccccccccccc}
\toprule
\multirow{2}{*}{System} & \multirow{2}{*}{Sensors} & \multirow{2}{*}{Algorithm} & \multicolumn{3}{c}{3D Volume} & \multicolumn{3}{c}{Z-MAP} & \multicolumn{2}{c}{Computation\tnote{a}} \\
\cmidrule(lr){4-6} \cmidrule(lr){7-9} \cmidrule(lr){10-11}
& & & PSNR$\uparrow$\tnote{b} & SSIM$\uparrow$ & MSE$\downarrow$ & PSNR$\uparrow$ & SSIM$\uparrow$ & MSE$\downarrow$ & Time$\downarrow$\tnote{c} & Speed \\
\midrule
\multirow{9}{*}{Planar} & \multirow{3}{*}{64}
& UBP & 24.78 & 0.2425 & 0.0033 & 12.83 & 0.0061 & 0.0522 & - & - \\
& & MB-PD & 26.27 & 0.5229 & 0.0024 & 14.62 & 0.0174 & 0.0345 & 1207.8 & - \\
& & \textbf{GPAIR} & \textbf{36.49}\tnote{d} & \textbf{0.9932} & \textbf{0.0002} & \textbf{18.67} & \textbf{0.7976} & \textbf{0.0136} & \textbf{3.92} & \textbf{308$\times$} \\
\cmidrule{2-11}
& \multirow{3}{*}{256}
& UBP & 25.65 & 0.2406 & 0.0027 & 14.49 & 0.0185 & 0.0356 & - & - \\
& & MB-PD & 29.53 & 0.5670 & 0.0011 & 17.26 & 0.0482 & 0.0188 & 4535.8 & - \\
& & \textbf{GPAIR} & \textbf{38.71} & \textbf{0.9970} & \textbf{0.0001} & \textbf{22.07} & \textbf{0.9183} & \textbf{0.0062} & \textbf{5.52} & \textbf{822$\times$} \\
\cmidrule{2-11}
& \multirow{3}{*}{1024}
& UBP & 27.20 & 0.2759 & 0.0019 & 16.71 & 0.0480 & 0.0213 & - & - \\
& & MB-PD & 33.93 & 0.7238 & 0.0004 & 19.33 & 0.1629 & 0.0117 & 17301.2 & - \\
& & \textbf{GPAIR} & \textbf{39.32} & \textbf{0.9985} & \textbf{0.0001} & \textbf{22.91} & \textbf{0.9635} & \textbf{0.0051} & \textbf{19.85} & \textbf{872$\times$} \\
\midrule
\multirow{9}{*}{\makecell{Hemi-\\spherical}} & \multirow{3}{*}{64}
& UBP & 24.24 & 0.1823 & 0.0038 & 12.95 & 0.0084 & 0.0506 & - & - \\
& & MB-PD & 26.85 & 0.5477 & 0.0021 & 15.09 & 0.0192 & 0.0310 & 985.0 & - \\
& & \textbf{GPAIR} & \textbf{36.96} & \textbf{0.9945} & \textbf{0.0002} & \textbf{19.29} & \textbf{0.8118} & \textbf{0.0118} & \textbf{4.46} & \textbf{221$\times$} \\
\cmidrule{2-11}
& \multirow{3}{*}{256}
& UBP & 25.06 & 0.1802 & 0.0031 & 14.88 & 0.0282 & 0.0325 & - & - \\
& & MB-PD & 30.19 & 0.6012 & 0.0010 & 17.65 & 0.0440 & 0.0172 & 3689.6 & - \\
& & \textbf{GPAIR} & \textbf{38.43} & \textbf{0.9982} & \textbf{0.0001} & \textbf{21.33} & \textbf{0.9411} & \textbf{0.0074} & \textbf{6.44} & \textbf{573$\times$} \\
\cmidrule{2-11}
& \multirow{3}{*}{1024}
& UBP & 28.46 & 0.2772 & 0.0014 & 18.57 & 0.0573 & 0.0139 & - & - \\
& & MB-PD & 34.57 & 0.7505 & 0.0003 & 19.37 & 0.1876 & 0.0116 & 13907.1 & - \\
& & \textbf{GPAIR} & \textbf{38.98} & \textbf{0.9987} & \textbf{0.0001} & \textbf{21.91} & \textbf{0.9598} & \textbf{0.0064} & \textbf{23.57} & \textbf{590$\times$} \\
\bottomrule
\end{tabular}
\label{tab:simulation_metrics}
\begin{tablenotes}
\item[a] All experiments used a reconstruction voxel grid size of $512 \times 512 \times 256 \approx 67$ million voxels.
\item[b] Unit: dB.
\item[c] Unit: s.
\item[d] Bold values indicate the best performance among compared methods.
\end{tablenotes}
\end{threeparttable}
\end{table}

\subsection{Validation on in vivo experimental results}

\begin{figure}[htbp]
\centering
\includegraphics[width=0.95\textwidth]{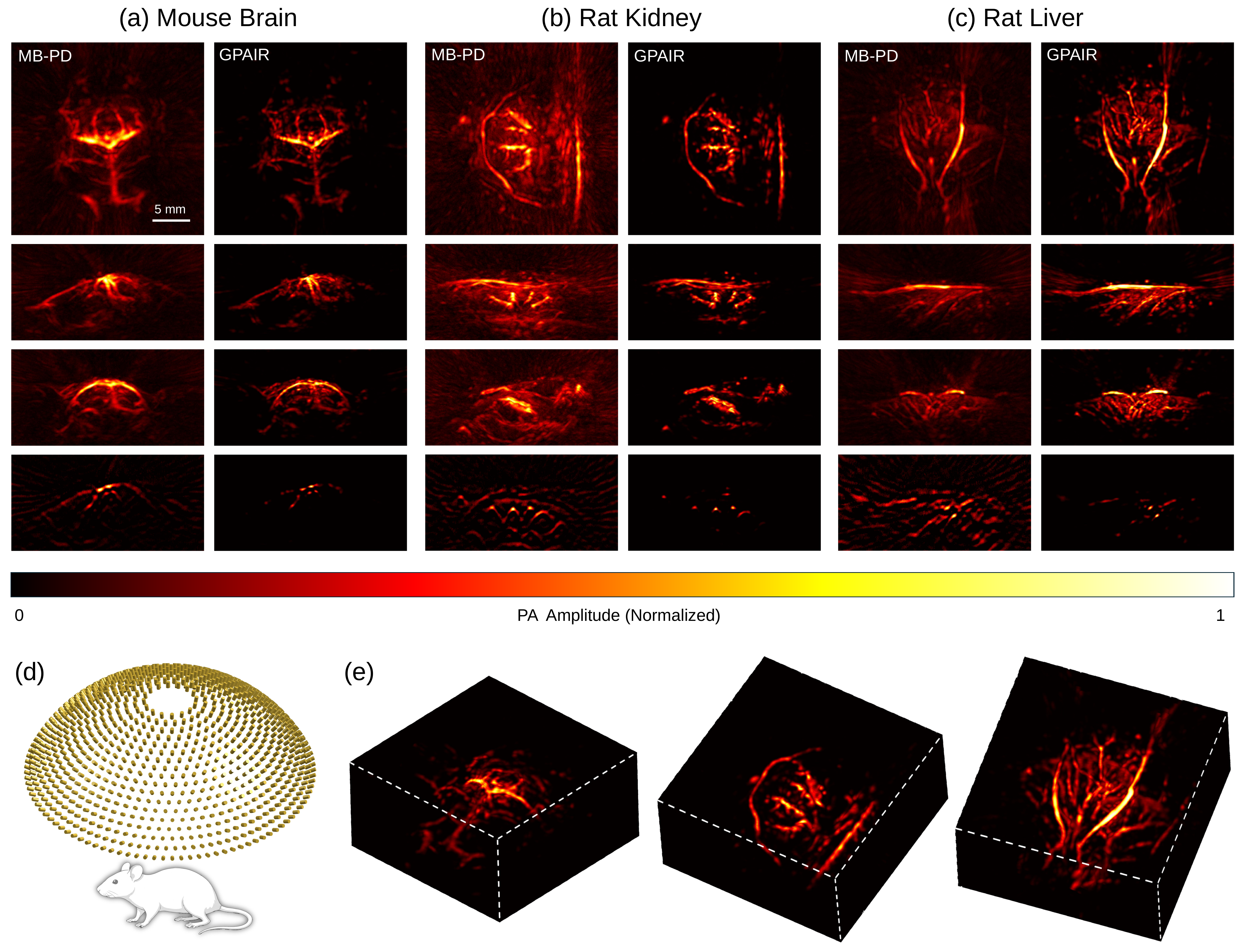}
\caption{In vivo 3D reconstruction results for mouse brain, rat kidney, and rat liver. (a)--(c) Reconstruction comparison between MB-PD and GPAIR for each organ. Rows: z-MAP, x-MAP, y-MAP, y-Slice. Scale bar: 5~mm. (d) Schematic of the experimental system with a 1024-element hemispherical detector array. (e) 3D volumetric rendering of GPAIR reconstructions for mouse brain, rat kidney, and rat liver.}
\label{fig:invivo}
\end{figure}

To validate GPAIR in realistic in vivo study, we processed in vivo data acquired using a high-performance 3D 1024-element hemispherical array from Professor Kim~\cite{choi2023deep, kim2022deep, kim20243d}. Three subjects were imaged: mouse brain, rat kidney and liver, with a reconstruction grid of $256 \times 256 \times 128$ voxels. Fig.~\ref{fig:invivo} compares the results from MB-PD and GPAIR.

For the mouse brain (Fig.~\ref{fig:invivo}(a)), the MAPs of GPAIR show reduced background noise compared to MB-PD. The vessel structures in the cortical region appear more continuous in the z-MAP. In the side views, GPAIR results exhibit less clutter between vascular layers. For the rat kidney (Fig.~\ref{fig:invivo}(b)), the slice view shows that GPAIR reconstructs vascular cross-sections with reduced artifacts, whereas MB-PD results appear more blurred with higher background levels. Similarly, in the rat liver (Fig.~\ref{fig:invivo}(c)), GPAIR recovers the hepatic vascular tree with higher contrast compared to MB-PD. The 3D volumetric renderings (Fig.~\ref{fig:invivo}(e)) visualize the 3D vascular networks of the three organs. Comprehensive comparisons including UBP results are provided in Supplementary Figures 4--6, Supplementary Note 2 and Supplementary Movies 4-6.

Quantitatively (Table~\ref{tab:invivo_metrics}), GPAIR achieved substantially higher contrast-to-noise ratios (CNR) compared to UBP and MB-PD. Background noise was consistently suppressed, and vessel sharpness was enhanced across all samples.Besides, in this typically practical study, GPAIR achieves sub-second reconstruction speed, while the compared IR method (MB-PD) still requires hundreds of seconds. To the best of our knowledge, this efficiency establishes GPAIR as the fastest IR method for 3D PACT reported to date, paving the way for high-fidelity real-time volumetric imaging.

\begin{table}[htbp]
\centering
\small
\setlength{\tabcolsep}{5pt}
\begin{threeparttable}
\caption{Quantitative Image Quality Metrics and Computational Performance Comparison in In Vivo Experiments}
\begin{tabular}{cccccccc}
\toprule
\multirow{2}{*}{Sample} & \multirow{2}{*}{Algorithm} & \multicolumn{4}{c}{Image Quality Metrics} & \multicolumn{2}{c}{Computation\tnote{a}} \\
\cmidrule(lr){3-6} \cmidrule(lr){7-8}
& & CNR$\uparrow$\tnote{b} & SNR$\uparrow$ & BG Std$\downarrow$ & Sharpness$\uparrow$ & Time$\downarrow$\tnote{c} & Speed \\
\midrule
\multirow{3}{*}{Mouse Brain}
& UBP   & 52.38 & 56.89 & 0.0146 & 0.88 & - & - \\
& MB-PD & 68.47 & 72.93 & 0.0119 & 0.81 & 237.2 & - \\
& \textbf{GPAIR} & \textbf{220.61}\tnote{d} & \textbf{290.93} & \textbf{0.0027} & \textbf{1.13} & \textbf{0.82} & \textbf{289$\times$} \\
\midrule
\multirow{3}{*}{Rat Kidney}
& UBP   & 29.23 & 39.43 & 0.0217 & 1.10 & - & - \\
& MB-PD & 46.87 & 59.14 & 0.0149 & 1.15 & 238.9 & - \\
& \textbf{GPAIR} & \textbf{871.08} & \textbf{1061.10} & \textbf{0.0007} & \textbf{1.47} & \textbf{0.80} & \textbf{299$\times$} \\
\midrule
\multirow{3}{*}{Rat Liver}
& UBP   & 52.14 & 56.80 & 0.0144 & 1.34 & - & - \\
& MB-PD & 85.25 & 81.30 & 0.0098 & 1.27 & 251.2 & - \\
& \textbf{GPAIR} & \textbf{203.30} & \textbf{99.19} & \textbf{0.0094} & \textbf{1.51} & \textbf{0.79} & \textbf{318$\times$} \\
\bottomrule
\end{tabular}
\label{tab:invivo_metrics}
\begin{tablenotes}
\footnotesize
\item[a] All experiments used a reconstruction voxel grid size of $256 \times 256 \times 128 \approx 8.4$ million voxels with 1024 detectors.
\item[b] Abbreviations: CNR, Contrast-to-Noise Ratio; SNR, Signal-to-Noise Ratio; BG Std, Background Standard Deviation.
\item[c] Unit: s.
\item[d] Bold values indicate the best performance among compared methods.
\end{tablenotes}
\end{threeparttable}
\end{table}

\subsection{Computational performance analysis}

\begin{figure}[htbp]
\centering
\includegraphics[width=0.95\textwidth]{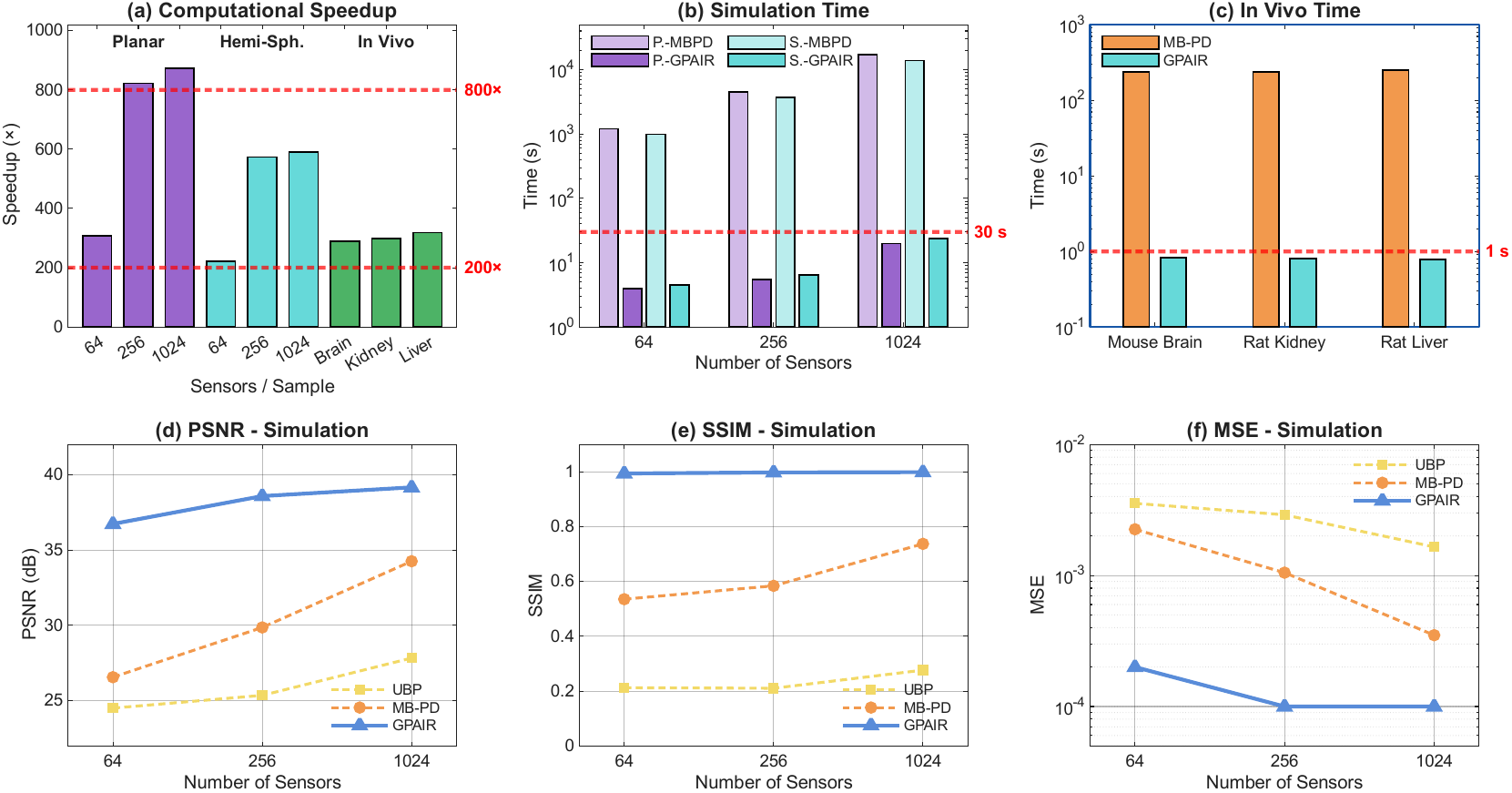}
\caption{Quantitative performance metrics comparison between GPAIR and other methods. (a) Computational speedup factors for planar simulations, hemispherical simulations, and in vivo experiments. Dashed lines indicate 200$\times$ and 800$\times$ levels. (b) Absolute reconstruction time for simulation experiments (log scale), with 30~s threshold marked. (c) Absolute reconstruction time for in vivo experiments (log scale), with 1~s threshold marked. (d)--(f) Quantitative image quality metrics (PSNR, SSIM, and MSE) for simulation experiments (Planar array) under varying sensor numbers.}
\label{fig:metrics}
\end{figure}

Beyond high image quality, GPAIR demonstrated revolutionary ultrafast computational speed. The quantitative performance is summarized in Fig.~\ref{fig:metrics}. As shown in Fig.~\ref{fig:metrics}(a), GPAIR consistently achieved orders-of-magnitude acceleration, with speedup factors ranging from over 200$\times$ to nearly 900$\times$ depending on the scale of the target and array configuration.

Fig.~\ref{fig:metrics}(b) and (c) detail the computation times on a logarithmic scale. In large-scale simulations (Fig.~\ref{fig:metrics}(b)), GPAIR completed all reconstructions within 30 seconds (indicated by the red dashed line), whereas MB-PD required up to several hours. Most remarkably, for in vivo datasets (Fig.~\ref{fig:metrics}(c)), GPAIR achieved sub-second reconstruction ($<0.82$~s) for 8.4-million-voxel volumes. This break-through to sub-second reconstruction fundamentally transforms 3D PACT from a tedious offline post-processing workflow into a near-real-time imaging display. Detailed iteration counts and timing comparisons are provided in Supplementary Note 8, with GPU memory consumption analysis in Supplementary Note 9.

\newpage

\section{Discussion}

This work presents GPAIR, a reconstruction framework that resolves the long-standing trade-off between modeling accuracy and computational efficiency in large-scale 3D PACT. By transitioning from discrete voxel-based approximations to a continuous Gaussian-kernel representation with analytical solutions, as well as techniques to take full advantages of GPU computing. GPAIR enables near-real-time volumetric photoacoustic image reconstruction with high quality.

The primary innovation of GPAIR lies in its departure from the discrete-to-discrete paradigm. Conventional methods approximate continuous biological tissues as piecewise cubic voxels, inevitably introducing discrete error. Although using denser spatial and temporal grids can effectively reduce errors, it will significantly sacrifice computing efficiency. In contrast, GPAIR models the initial pressure field as a superposition of continuous isotropic Gaussian kernels, ensuring spatial differentiability. The closed-form analytical solution for the spherical integral eliminates temporal time-stepping and look-up table interpolation, thereby bypassing numerical errors inherent to discrete approximations.

The orders' of acceleration in computation represents a transformative leap for 3D PACT. This acceleration stems from both the analytical formulation and hardware-aware implementation. By utilizing GPU-native Triton kernels with optimized memory access patterns, GPAIR maximizes throughput on modern architectures, overcoming the memory bandwidth limitations that typically bottleneck iterative algorithms. Importantly, this efficiency provides substantial computational headroom: under extreme sparse sampling conditions, we could double the iteration count to ensure robust convergence while maintaining reconstruction times well under practical thresholds. Such flexibility enables the integration of advanced regularization priors without compromising real-time feasibility.

Beyond the complete iterative reconstruction framework, the core computational modules of GPAIR, the forward operator and adjoint operator, have been extracted as standalone high-performance tools called SignalSimulator and ImageReconstructor. The SignalSimulator achieves superior fast forward projection speed of 37.41~ms for in vivo configurations with 8.4 million voxels with 1024 detectors, representing approximately three orders of magnitude speedup compared to k-Wave (Supplementary Figure 7 and Supplementary Notes 3). The ImageReconstructor achieves a remarkable 4.99~ms for in vivo data with also 8.4 million voxels and 1024 detectors, enabling over 200 reconstructions per second (Supplementary Figures 8--9 and Supplementary Notes 4--5). This modular design allows researchers to use the SignalSimulator for rapid forward modeling or deep learning data augmentation, and employ the ImageReconstructor for real-time visualization or as initialization for iterative algorithms.

The robustness of GPAIR under limited-view and sparse sampling conditions has demonstrated the efficacy of the proposed vessel continuity regularization (VCR). Standard total variation regularization, while effective at noise suppression, tends to promote piecewise constant structures, causing ``staircasing'' artifacts at smooth vessel boundaries. By incorporating the Hessian-based term that penalizes second-order derivatives, GPAIR specifically preserves the geometric curvature characteristic of tubular structures. This composite prior effectively connects fragmented vessel segments and maintains boundary smoothness even with limited angular coverage. Notably, GPAIR successfully recovered vertical vessel segments in planar array geometries, which are typically hard to view to conventional methods due to limited detection aperture, by exploiting high-frequency temporal information combined with spatial continuity constraints. Comprehensive robustness analyses demonstrate that GPAIR maintains excellent reconstruction quality under varying noise levels up to SNR $\approx$ 5:1 (Supplementary Figure 10 and Supplementary Note 6), and reveals that angular coverage is more critical than sampling density for sparse sampling scenarios (Supplementary Figure 11 and Supplementary Note 7).

The transition to sub-second reconstruction holds profound clinical implications. Currently, 3D PACT is largely confined to offline post-processing workflows. GPAIR closes this latency gap, enabling visualization at speeds compatible with intraoperative navigation and microsurgical guidance. For dynamic functional imaging applications, such as pharmacokinetic tracking or hemodynamic monitoring, the preserved temporal resolution facilitates high-fidelity 4D PACT (3D space + time), potentially enabling new quantitative metrics for vascular physiology without motion artifacts or temporal blurring.

Several directions remain for further refinement. First, the uniform speed of sound assumption may introduce aberrations in acoustically heterogeneous environments; future work could incorporate differentiable ray-tracing for phase aberration correction. Second, isotropic kernels with fixed standard deviations may be suboptimal for highly anisotropic structures; extending the model to learnable anisotropic 3D Gaussians could enhance resolution and sparsity. Finally, as a differentiable physics layer, GPAIR can be integrated into deep learning pipelines, where unrolled network architectures with learned data-driven priors may further advance reconstruction performance.

In conclusion, GPAIR establishes a new benchmark for ultrafast 3D photoacoustic image reconstruction. By synergizing continuous physics modeling with high-performance GPU computing, GPAIR achieves orders-of-magnitude acceleration while delivering superior image quality. This advancement transforms 3D PACT from a computationally prohibitive research tool into a viable near-real-time clinical imaging modality, poised to advance the diagnosis and monitoring of vascular diseases.

\newpage

\section{Methods}

\subsection{Gaussian-kernel-based distribution for the continuous source discretization}

We represent the initial pressure field as a linear superposition of isotropic Gaussian kernels centered at spatial griding points, as shown in Fig. \ref{fig:methods}(a). Specifically, assuming there are $M$ voxels in the reconstruction grid, denoted as $i=1,2,3,\dots,M$. Each voxel $i$ is assigned as an isotropic 3D Gaussian kernel centered at the voxel center $\mathbf{r}_i$ with an initial pressure $A_i$ and a standard deviation $\sigma$, described as below:

\begin{equation}
{p_0}_i(\mathbf{r}) = A_i  e^{-\frac{\|\mathbf{r} - \mathbf{r}_i\|^2}{2\sigma^2}}.
\end{equation}

Then, for arbitrary position $\mathbf{r}$, its initial pressure $p_0(\mathbf{r})$ is the superposition of pressures contributed from all Gaussian kernels:

\begin{equation}
p_0(\mathbf{r}) = \sum_{i=1}^{M} {p_0}_i(\mathbf{r}) = \sum_{i=1}^{M} A_i e^{-\frac{\|\mathbf{r} - \mathbf{r}_i\|^2}{2\sigma^2}}.
\end{equation}

According to the wave property, the PA pressure wave detected by an ultrasound detector is also the linear superposition of waves generated from all Gaussian kernel sources.

Now we derive pressure wave generated by a single Gaussian kernel source. To simplify the derivation, we place the detector under consideration at the coordinate origin. Consider a Gaussian kernel source centered at $\mathbf{r}$ with amplitude $A$ and standard deviation $\sigma$. Its PA pressure received by the detector at time $t$ is~\cite{xu2005universal}:

\begin{equation}
p(\mathbf{r},t)=\frac{1}{4\pi v}\frac{\partial}{\partial t}\left(\frac{1}{vt}\int_{S^\prime(t)}Ae^{-\frac{\|\boldsymbol{\boldsymbol{\mathrm{r}-\mathrm{r}}^\prime}\|^2}{2\sigma^2}}\mathrm{d}S^\prime\right).
\label{eq:pa_wave}
\end{equation}

Where $v$ is the speed of sound, $S^\prime(t)$ is the intersection surface between the Gaussian kernel and the sphere centered at the detector with radius $r' = vt$ at time $t$, and $\mathbf{r}^\prime$ is the vector that points to the intersection surface, as shown in Fig. \ref{fig:methods}(c). We denote

\begin{align}
I(\mathbf{r},t)&=\int_{S^\prime(t)}Ae^{-\frac{\|\mathbf{r}-\mathbf{r}^\prime\|^2}{2\sigma^2}}\mathrm{d}S^\prime,\quad r^\prime=vt.
\end{align}:

Remarkably, the above spherical surface integral $I$ has a closed-form analytical solution:

\begin{align}
I(r,t)=A\frac{2\pi\sigma^2 r^\prime}{r}\left(e^{-\frac{(r-r^\prime)^2}{2\sigma^2}}-e^{-\frac{(r+r^\prime)^2}{2\sigma^2}}\right),\quad r^\prime=vt.
\label{eq:sphere_int}
\end{align}

Then, substituting Eq. (\ref{eq:sphere_int}) into Eq. (\ref{eq:pa_wave}), we have:

\begin{align}
\begin{split}
p(r,t)&=\frac{1}{4\pi v}\frac{\partial}{\partial t}\left(\frac{1}{vt}I\right)
=\frac{1}{4\pi}\frac{\partial}{\partial r^\prime}\left(\frac{I}{r^\prime}\right)
=\frac{1}{4\pi{r^\prime}^2}\left(r^\prime\frac{\partial I}{\partial r^\prime}-I\right)\\
\quad\\
&=\frac{A}{2r}\left((r-r^\prime)e^{-\frac{(r-r^\prime)^2}{2\sigma^2}}+(r+r^\prime)e^{-\frac{(r+r^\prime)^2}{2\sigma^2}}\right)\\
\quad\\
&=\frac{A}{2r}\left((r-vt)e^{-\frac{(r-vt)^2}{2\sigma^2}}+(r+vt)e^{-\frac{(r+vt)^2}{2\sigma^2}}\right).
\end{split}
\end{align}

The derived pressure expression consists of two terms with distinct physical meanings. The first term, $\frac{A}{2r}(r-vt)e^{-\frac{(r-vt)^2}{2\sigma^2}}$, represents the outgoing spherical wave propagating from the source. This is the primary signal observed in experiments and exhibits a characteristic N-shaped-like profile, as shown in Fig. \ref{fig:methods}(d). The second term, $\frac{A}{2r}(r+vt)e^{-\frac{(r+vt)^2}{2\sigma^2}}$, corresponds to an incoming wave (or inward-propagating wave) converging towards the source center. To guarantee spatial resolution, the standard deviation $\sigma$ should be smaller than the expected resolution. In this study, we set it equal to the grid size ($\sigma=\Delta x$). For $t>0$ and under typical far-field detection conditions (where the source-to-detector distance $r \gg \sigma$), the magnitude of the second term vanishes due to rapid exponential decay. Consequently, we neglect this component in practical photoacoustic imaging applications.

Therefore, the above formulas can be rewritten. Let $d=r-vt$ denote the distance from the integration surface to the center of the Gaussian source at time $t$, we have the forward process:

\begin{align}
\begin{split}
p(r,t)&=\frac{A}{2r}(r-vt)e^{-\frac{(r-vt)^2}{2\sigma^2}}\\
\quad\\
&=\frac{A}{2r}de^{-\frac{d^2}{2\sigma^2}}.
\end{split}
\label{eq:single_forward}
\end{align}

In practice, considering the $3\sigma$ principle, we set $-3\sigma<d<3\sigma$. This representation has the following advantages: the smoothness of Gaussian kernels ensures that the pressure field is continuously differentiable in space, avoiding discontinuities at voxel boundaries; the exponential decay property of Gaussian kernels ensures that each voxel's influence is limited, avoiding global coupling; the analytical properties of Gaussian functions allow both the forward model and its gradient to be expressed in closed form, enabling fast computing and gradient-based optimization; due to the linearity of the wave equation, the detected PA pressure is the linear superposition of multiple Gaussian sources, which is the core of subsequent efficient algorithm design.

\subsection{Adaptive supersampling alignment for the differentiable model operator}

With the analytical expression for a single Gaussian source derived in Eq. (\ref{eq:single_forward}), a straightforward approach to compute the forward signal is to iterate over all source-detector pairs and evaluate the contribution at each temporal sampling point where $d$ falls within the $\pm 3\sigma$ range. While this approach is mathematically rigorous, its computational complexity is prohibitive: for $M$ voxels and $N_d$ detectors, each requiring evaluation at multiple time points, the total computation scales as $\mathcal{O}(M N_d N_\sigma)$, where $N_\sigma$ denotes the number of samples spanning the Gaussian kernel support. This direct enumeration becomes intractable for large-scale 3D reconstructions.

To achieve ultrafast computation, we exploit the shift-invariance property of the Gaussian-derivative kernel by aligning the time-of-flight (ToF) from each source center to each detector onto discrete temporal grid points~\cite{ding2020model}. This alignment enables the use of highly optimized 1D transposed convolution operations, which efficiently ``scatter'' the kernel waveform at each aligned impulse location. Specifically, once all ToFs are snapped to grid indices, the forward signal synthesis reduces to a single transposed convolution operation that can be massively parallelized on GPUs, reducing the complexity to near-linear scaling with the number of non-zero impulses.

However, aligning the continuous ToF to the nearest grid point on the original sampling grid introduces temporal quantization errors. When mapped back to the spatial domain, these errors manifest as jitter artifacts in the reconstruction. To reconcile the conflict between efficient shift-invariant convolution and sub-grid ToF precision, we propose the adaptive supersampling alignment (ASSA) scheme. The key insight is to first upsample the temporal grid to a finer resolution, thereby reducing the maximum alignment error to a negligible fraction of the target spatial resolution. The forward signal is synthesized on this upsampled grid and then downsampled back to the original sampling rate for loss computation and iterative optimization. Crucially, all operations in this pipeline, including sparse projection, transposed convolution, and decimation, remain linear operators. This linearity not only enables the rigorous derivation of the adjoint operator $\mathcal{A}^T$ for gradient-based optimization but also ensures that the entire forward-backward pass can be efficiently implemented using standard tensor operations, thereby achieving both computational efficiency and optimization tractability.

\subsubsection*{Adaptive Grid Parameters}

Let $\mathbf{y} \in \mathbb{R}^{N_d \times N_t}$ denote the discretized pressure signals, where $y_j[n]$ represents the signal at the $j$-th detector and time index $n$. Since the temporal width of the Gaussian kernel is determined by the acoustic transit time across its support (approximately $3\sigma/v_s$), we determine the adaptive upsampling ratio $\alpha$ by constraining the minimum number of sampling points covering the kernel. This effectively reduces the maximum temporal alignment error (half the upsampled interval, $\Delta t^{\text{up}}/2$). The adaptive upsampling ratio $\alpha$ is calculated as:

\begin{align}
\begin{split}
N_{\text{half}} &= \left\lceil \frac{3\sigma}{v_s \Delta t} \right\rceil, \quad \alpha = \max\left(1, \left\lceil \frac{(N_{\min}-1)/2}{N_{\text{half}}} \right\rceil \right),\\
\quad\\
f_s^{\text{up}} &= \alpha f_s, \quad \Delta t^{\text{up}} = \frac{1}{f_s^{\text{up}}}, \quad N_t^{\text{up}} = \alpha N_t.
\end{split}
\end{align}

Here, $N_{\text{half}}$ is the number of original sampling points within the kernel's half-width, and $N_{\min}$ is the target minimum points, which is set to 25 in this work, to ensure smoothness. Then we can define a high-density intermediate grid with $N_t^{\text{up}}$ points.

\subsubsection*{Forward Discrete Operator}

Based on this high-density grid, we rigorously define the forward operator $\mathcal{A}$. The process consists of three distinct stages: projection to the upsampled temporal grid, kernel convolution, and downsampling.

First, let $\mathbf{z} \in \mathbb{R}^{N_d \times N_t^{\text{up}}}$ be the intermediate sparse impulse sequence. We define the discretized source distribution as a vector $\mathbf{x} \in \mathbb{R}^M$, where $M$ denotes the total number of spatial voxels. For a voxel $i$ with initial pressure $x_i$ at position $\mathbf{r}_i$ and a detector $j$ at $\mathbf{s}_j$, the time-of-flight is aligned to the nearest integer index $k_{ij}$ on the upsampled grid. This operation defines the projection operator $\mathcal{P}_{\text{up}}$:

\begin{align}
\begin{split}
r_{ij} &= \|\mathbf{r}_i - \mathbf{s}_j\|,\quad k_{ij} = \left\lfloor \frac{r_{ij}}{v_s} \cdot f_s^{\text{up}} + 0.5 \right\rfloor,\\
\quad\\
z_j[k] &= \sum_{i=1}^{M} \frac{x_i}{r_{ij}} \cdot \mathbbm{1}(k = k_{ij}), \quad \text{for } k=0,\ldots,N_t^{\text{up}}-1.
\end{split}
\end{align}

Where $\mathbbm{1}(\cdot)$ is the indicator function. This effectively ``snaps'' the continuous PA pulse center $k_{ij}$ from each Gaussian kernel to the grid, enabling the use of the convolution kernel later.

Next, the upsampled forward model waveform $\tilde{\mathbf{z}} \in \mathbb{R}^{N_d \times N_t^{\text{up}}}$ is synthesized by applying a transposed convolution with a kernel $h$ to the sparse temporal sequence $\mathbf{z}$:

\begin{equation}
\tilde{\mathbf{z}} = h * \mathbf{z}, \quad \tilde{z}_j[k] = (h * z_j)[k] = \sum_{m=0}^{N_t^{\text{up}}-1} z_j[m] \cdot h[k - m].
\end{equation}

Here, the kernel $h$ is defined as (according to Eq. (\ref{eq:single_forward})):

\begin{equation}
h[k] = C \cdot d[k] \cdot e^{-\frac{d[k]^2}{2\sigma^2}}, \quad\text{where } d[k] = -v_s k \Delta t^{\text{up}}, \quad k = -K, \ldots, K,
\end{equation}

with $K = \alpha N_{\text{half}}$ being the upsampled kernel's half-width and $C$ a normalization constant. The transposed convolution effectively ``scatters'' the kernel $h$ at each non-zero impulse location, which can be implemented via high-efficient 1D transposed convolution operator in modern GPU computing architecture.

Finally, a decimation operator $\mathcal{S}_{\downarrow \alpha}$ restores the signal back to the original sampling rate via sampling every each $\alpha$ points:

\begin{equation}
y_j[n] = (\mathcal{S}_{\downarrow \alpha} \tilde{z}_j)[n] = \tilde{z}_j[\alpha \cdot n], \quad n=0,\ldots,N_t-1.
\end{equation}

The complete forward model is compactly expressed as:

\begin{equation}
\mathbf{y} = \mathcal{A}(\mathbf{x}) = \mathcal{S}_{\downarrow \alpha} \left( h * \mathcal{P}_{\text{up}}(\mathbf{x}) \right).
\end{equation}

\subsubsection*{Adjoint Operator}

The adjoint operator $\mathcal{A}^T$, required for gradient-based reconstruction, is derived by the transpose of each stage.

\begin{equation}
\mathbf{g} = \mathcal{A}^T(\boldsymbol{\delta}) = \mathcal{P}_{\text{up}}^T \left( \bar{h} \ast \mathcal{S}_{\downarrow \alpha}^T(\boldsymbol{\delta}) \right).
\end{equation}

Here, $\boldsymbol{\delta} \in \mathbb{R}^{N_d \times N_t}$ represents the residual or gradient at the detectors, and $\bar{h}$ is the time-reversed kernel defined as $\bar{h}[k] = h[-k]$. The components function as follows:

\begin{enumerate}
    \item Transpose Decimation (Zero-filling): $\mathcal{S}_{\downarrow \alpha}^T$ upsamples the residual by inserting zeros between sampling points:
    \begin{equation}
        \delta^{\text{up}}_j[k] = 
        \begin{cases} 
        \delta_j[k/\alpha] & \text{if } k \pmod \alpha = 0 \\
        0 & \text{otherwise}
        \end{cases},\quad k=0,\ldots,N_t^{\text{up}}-1.
    \end{equation}
    
    \item Adjoint of Transposed Convolution: The adjoint of the transposed convolution is the standard convolution $*$ with the time-reversed kernel $\bar{h}[k]$:
    \begin{equation}
        \delta^{\text{conv}}_j[k] = (\bar{h} \ast \delta^{\text{up}}_j)[k] = \sum_{m} \bar{h}[k-m] \delta^{\text{up}}_j[m] = \sum_{m} h[m-k] \delta^{\text{up}}_j[m].
    \end{equation}
    Note that, the derivative kernel $h$, is an odd function (anti-symmetric), $\bar{h}[k] = h[-k] = -h[k]$.
    
    \item Transpose Projection (Back-projection): $\mathcal{P}_{\text{up}}^T$ accumulates the gradients by sampling the correlated signal at each aligned index $k_{ij}$:
    \begin{equation}
        g_i = \sum_{j=1}^{N_d} \frac{\delta^{\text{conv}}_j[k_{ij}]}{r_{ij}}.
    \end{equation}
\end{enumerate}

The implementation details are summarized in Algorithm \ref{alg:forward_adjoint}. In summary, here we address the critical challenge of reconciling continuous physical ToF with discrete computational grids. The proposed ASSA scheme adaptively determines the upsampling ratio based on the spatial resolution requirements, ensuring that temporal quantization errors remain negligible relative to the target imaging resolution. By decomposing the forward operator into three cascaded stages: sparse projection, shift-invariant convolution, and decimation, we enable the use of highly optimized transposed convolution primitives while preserving sub-grid temporal accuracy. The corresponding adjoint operator is rigorously derived to ensure mathematical consistency, which is essential for gradient-based optimization. This formulation transforms the computationally intensive forward modeling into a series of parallelizable operations amenable to GPU acceleration, laying the algorithmic foundation for the subsequent iterative reconstruction framework.

\begin{algorithm}[htbp]
\caption{Forward and Adjoint Operators with ASSA}
\label{alg:forward_adjoint}
\begin{algorithmic}[1]
\Require Precomputed: lookup table $\{r_{ij}, k_{ij}\}$, kernel $h[k] \in \mathbb{R}^{2K+1}$, ratio $\alpha$
\Ensure Forward: $\mathbf{y} = \mathcal{A}(\mathbf{x})$; Adjoint: $\mathbf{g} = \mathcal{A}^T(\boldsymbol{\delta})$
\State
\Function{ForwardOp}{$\mathbf{x} \in \mathbb{R}^M$}
    \State // \textit{1. Project to sparse upsampled grid} ($\mathcal{P}_{\text{up}}$)
    \State Initialize sparse buffer $\mathbf{z} \gets \mathbf{0} \in \mathbb{R}^{N_d \times \alpha N_t}$
    \For{$i = 1, \ldots, M$; $j = 1, \ldots, N_d$}
        \State $\mathbf{z}_j[k_{ij}] \mathrel{+}= x_i / r_{ij}$  \Comment{Accumulate masses at aligned indices}
    \EndFor
    \State // \textit{2. Kernel scattering via transposed convolution}
    \State $\tilde{\mathbf{z}} \gets h * \mathbf{z}$ \Comment{Transposed convolution on sparse grid}
    \State // \textit{3. Decimation} ($\mathcal{S}_{\downarrow \alpha}$)
    \State $\mathbf{y}[n] \gets \tilde{\mathbf{z}}[\alpha \cdot n]$, $\forall n = 0, \ldots, N_t-1$
    \State \Return $\mathbf{y}$
\EndFunction
\State
\Function{AdjointOp}{$\boldsymbol{\delta} \in \mathbb{R}^{N_d \times N_t}$}
    \State // \textit{1. Zero-filling upsampling} ($\mathcal{S}_{\downarrow \alpha}^T$)
    \State Initialize $\boldsymbol{\delta}^{\text{up}} \gets \mathbf{0} \in \mathbb{R}^{N_d \times \alpha N_t}$
    \State $\boldsymbol{\delta}^{\text{up}}[\alpha \cdot n] \gets \boldsymbol{\delta}[n]$, $\forall n$
    \State // \textit{2. Standard convolution (Adjoint of transposed convolution)}
    \State Obtain flipped kernel $\bar{h}[k] \gets h[-k]$
    \State $\boldsymbol{\delta}^{\text{conv}} \gets \bar{h} \ast \boldsymbol{\delta}^{\text{up}}$
    \State // \textit{3. Back-projection from specific indices} ($\mathcal{P}_{\text{up}}^T$)
    \State Initialize $\mathbf{g} \gets \mathbf{0} \in \mathbb{R}^M$
    \For{$i = 1, \ldots, M$; $j = 1, \ldots, N_d$}
        \State $g_i \mathrel{+}= \boldsymbol{\delta}_j^{\text{conv}}[k_{ij}] / r_{ij}$
    \EndFor
    \State \Return $\mathbf{g}$
\EndFunction
\end{algorithmic}
\end{algorithm}

\subsection{Gradient-based reconstruction framework with physical constraints and anatomical priors}

Leveraging the differentiable forward operator $\mathcal{A}$ and its adjoint $\mathcal{A}^T$ derived in the above subsection, we formulate the photoacoustic image reconstruction as a large-scale inverse optimization problem. Unlike traditional IR methods that often rely on discrete ray-tracing approximations, our framework utilizes the analytical differentiability of the Gaussian-kernel-based model to perform exact gradient descent. This allows us to recover the initial pressure distribution $\mathbf{x}$ by minimizing the discrepancy between the simulated data $\mathcal{A}(\mathbf{x})$ and the detected PA signals $\mathbf{b}$. To robustly solve this ill-posed inverse problem, particularly under sparse-view detection scenarios, we propose a comprehensive optimization framework that incorporates high-performance computing implementation, strict physical constraints, and anatomical regularization priors.

\subsubsection*{High-performance differentiable implementation}

The computational bottleneck of IR lies in the repeated execution of the forward and adjoint operators. The back-projection step in the adjoint operator $\mathcal{A}^T$, specifically, involves scattering gradients from detector time points to voxel locations, which requires massive random memory access and atomic accumulation operations. Conventional CPU-based implementations typically require minutes to hours for large-scale 3D reconstructions, rendering iterative methods impractical for clinical applications.

To address this, we implemented both $\mathcal{A}$ and $\mathcal{A}^T$ as custom autograd functions using Triton~\cite{tillet2019triton}, a domain-specific language for block-oriented GPU programming. Our implementation incorporates several key optimizations: (1) Structure of Arrays (SoA) memory layout to maximize coalesced memory access and minimize cache misses; (2) Split-K parallelization strategy that distributes the detector-voxel pair computations across multiple thread blocks, enabling efficient load balancing; and (3) hardware-accelerated atomic operations with carefully designed blocked memory access patterns that align with the GPU memory hierarchy. These optimizations ensure that the gradient computation is both numerically precise and computationally efficient.

Benchmark on an RTX 5090 GPU demonstrates that our implementation achieves 37.41 ms for forward projection and 4.99 ms for back-projection with 8.4 million voxels. Detailed performance is provided in Supplementary Notes 3--5. This end-to-end differentiable implementation allows us to seamlessly integrate the physics-based model into powerful modern optimization frameworks used in deep learning (e.g., PyTorch~\cite{paszke2019pytorch}).

\subsubsection*{Nonnegative parameterization constraint}

The PA initial pressure $\mathbf{x}$, representing the deposited optical energy density, is physically constrained to be non-negative. Standard constrained optimization methods, such as projected gradient descent, enforce this by truncating negative values at each iteration. However, this ``hard clipping'' introduces discontinuities in the optimization landscape and can disrupt the gradient flow, potentially leading to suboptimal convergence.
Instead, we adopt a nonnegative parameterization constraint (NPC) strategy. We reparameterize the initial pressure $\mathbf{x}$ using an unconstrained latent variable $\mathbf{z} \in \mathbb{R}^M$:

\begin{equation}
\mathbf{x} = \phi(\mathbf{z}) = (\mathbf{z} + \epsilon)^2.
\end{equation}

Where the element-wise squaring ensures non-negativity naturally, and $\epsilon = 10^{-8}$ is a small constant added to prevent gradient vanishing when $\mathbf{z}$ approaches zero. This transformation converts the constrained problem into an unconstrained one with respect to $\mathbf{z}$, allowing the use of standard unconstrained optimizers. The gradient is backpropagated via the chain rule:

\begin{equation}
\frac{\partial \mathcal{L}}{\partial \mathbf{z}} = \frac{\partial \mathcal{L}}{\partial \mathbf{x}} \odot 2(\mathbf{z} + \epsilon).
\end{equation}

Where $\odot$ denotes the Hadamard product.

\subsubsection*{Vessel continuity regularization}

In vascular photoacoustic imaging, the reconstruction is often underdetermined due to limited detector coverage. Pure data-fidelity optimization tends to produce artifacts such as streak noise or fragmented vessels. Conventional regularization approaches exhibit limitations for vascular imaging: Tikhonov ($L_2$) regularization~\cite{golub1999tikhonov} over-smoothes vessel boundaries; standard total variation (TV) regularization~\cite{chambolle2004algorithm} produces staircase artifacts along elongated structures and fails to preserve vessel continuity; higher-order regularization alone may over-smooth directional variations.

To address these major limitations, we introduce the vessel continuity regularization (VCR), specifically designed to preserve the morphological characteristics of blood vessels. Blood vessels exhibit two key geometric features: (1) Tubular structure, where intensity varies smoothly along the vessel but changes rapidly in the cross-sectional direction; and (2) Piecewise smoothness, with uniform vessel interior and sharp boundaries. Accordingly, our VCR term combines Hessian-based regularization and TV regularization in a synergistic manner: the Hessian term enforces curvature continuity along vessel trajectories to connect fragmented segments, while the TV term preserves sharp vessel-background boundaries:

\begin{equation}
\mathcal{R}_{\text{VCR}}(\mathbf{x}) = \mathcal{R}_{\text{H}}(\mathbf{x}) + \beta \cdot \mathcal{R}_{\text{TV}}(\mathbf{x}).
\end{equation}

The Hessian regularization~\cite{lefkimmiatis2011hessian} penalizes the second-order derivatives of the image, which correspond to local curvature. By minimizing the nuclear norm or Frobenius norm of the Hessian matrix, we encourage the reconstruction to favor structures with bounded curvature, effectively connecting fragmented vessel segments. We implement this using a rotationally invariant second-order discrete difference operator $\mathcal{D}$:

\begin{equation}
\mathcal{R}_{\text{H}}(\mathbf{x}) = \sum_{i=1}^M \sqrt{ \sum_{p,q \in \{x,y,z\}} (\mathcal{D}_{pq} x_i)^2 + \epsilon }.
\end{equation}

Here, $\mathcal{D}_{pq}x_i$ is the second-order difference operator $\mathcal{D}_{pq}$ applied at voxel $i$.

The TV regularization complements the Hessian term by penalizing the first-order gradient magnitude. This promotes sparsity in the gradient domain, effectively suppressing background noise and sharpening vessel edges without blurring:

\begin{equation}
\mathcal{R}_{\text{TV}}(\mathbf{x}) = \sum_{i=1}^M \sqrt{ \sum_{d \in \{x,y,z\}} (\mathcal{D}_d x_i)^2 + \epsilon }.
\end{equation}

The hyperparameter $\beta$ balances the trade-off between structural continuity (Hessian) and edge sharpness (TV).

\subsubsection*{Optimization formulation and algorithm}

Integrating the data fidelity term, physical constraints, and regularization, the complete reconstruction problem is formulated as:

\begin{align}
\mathop{\arg\min}\limits_{\mathbf{z}}\ \mathcal{L}(\mathbf{z}) &= \mathcal{L}_{\text{data}} + \lambda \cdot \mathcal{R}_{\text{VCR}} = \frac{1}{N}\|\mathcal{A}(\phi(\mathbf{z})) - \mathbf{b}\|_2^2 + \lambda \cdot \mathcal{R}_{\text{VCR}}(\phi(\mathbf{z})).
\end{align}

Where $N = N_d \times N_t$ denotes the total number of data points, and $\lambda$ controls the global regularization strength. We employ the Adam optimizer~\cite{kingma2014adam} to solve this non-convex problem due to its adaptive momentum properties, which accelerate convergence in complex landscapes. To further avoid getting trapped in local minima, which is a common issue in high-dimensional iterative reconstruction, we utilize a cosine annealing warm restarts~\cite{loshchilov2016sgdr} learning rate scheduler as follows. This strategy periodically resets the learning rate, allowing the optimizer to explore different basins of attraction and ultimately converge to a more robust solution.

\begin{align}
\begin{split}
\eta_t &= \eta_{\min} + \frac{1}{2}(\eta_{\max} - \eta_{\min})\left(1 + \cos\left(\frac{T_{\text{cur}}}{T_i}\pi\right)\right),\\
\quad\\
T_{\text{cur}} &= t\ \ \text{mod}\ \ T_0, \quad T_i = T_0 \cdot T_{\text{mult}}^{\lfloor t / T_0 \rfloor}.
\end{split}
\end{align}

Where $\eta_{\max}$ is the initial learning rate, $\eta_{\min}$ is the minimum learning rate, $T_0$ is the first restart period, and $T_{\text{mult}}$ is the period multiplier. The complete reconstruction workflow, named Gaussian-Kernel-Based Ultrafast 3D Photoacoustic Iterative Reconstruction (GPAIR), is summarized in Algorithm \ref{alg:gpair}.

In conclusion, we present a comprehensive gradient-based reconstruction framework that synergistically integrates computational efficiency, physical validity, and anatomical plausibility. The high-performance implementation leveraging Triton-based GPU kernels ensures that the iterative optimization remains computationally tractable even for large-scale 3D reconstructions. The nonnegative parameterization constraint elegantly transforms the physically constrained problem into an unconstrained optimization landscape, preserving gradient continuity and facilitating stable convergence. Furthermore, the vessel continuity regularization, combining Hessian-based curvature penalty with total variation, effectively suppresses reconstruction artifacts while preserving the tubular morphology characteristic of vascular structures. Collectively, these methodological components constitute the GPAIR algorithm, which achieves a principled balance between data fidelity and prior knowledge, enabling high-quality photoacoustic image reconstruction under challenging sparse-view acquisition scenarios.

\begin{algorithm}[htbp]
\caption{GPAIR: Gaussian-Kernel-Based Ultrafast 3D Reconstruction}
\label{alg:gpair}
\begin{algorithmic}[1]
\Require Observed data $\mathbf{b}$, System matrix operators $\mathcal{A}, \mathcal{A}^T$ (implemented via Algorithm \ref{alg:forward_adjoint})
\Require Hyperparameters: steps $I_{\max}$, learning rate range $[\eta_{\min}, \eta_{\max}]$, regularization weights $\lambda, \beta$
\Ensure Reconstructed image $\mathbf{x}^*$
\State
\State \textbf{1. Initialization}
\State $\mathbf{z}^{(0)} \leftarrow 0$ \Comment{Zero initialization}
\State Initialize Adam optimizer state $\mathbf{m} \leftarrow 0, \mathbf{v} \leftarrow 0$
\State
\State \textbf{2. Iterative Optimization}
\For{$t = 0$ to $I_{\max}-1$}
    \State \textit{// Forward Pass}
    \State $\mathbf{x}^{(t)} \leftarrow (\mathbf{z}^{(t)} + \epsilon)^2$ \Comment{Apply NPC}
    \State $\mathbf{y}^{(t)} \leftarrow \mathcal{A}(\mathbf{x}^{(t)})$ \Comment{Algorithm \ref{alg:forward_adjoint}: ForwardOp}
    \State
    \State \textit{// Loss Calculation}
    \State $\mathcal{L}_{\text{fidelity}} \leftarrow \frac{1}{N}\|\mathbf{y}^{(t)} - \mathbf{b}\|_2^2$
    \State $\mathcal{L}_{\text{reg}} \leftarrow \mathcal{R}_{\text{H}}(\mathbf{x}^{(t)}) + \beta \mathcal{R}_{\text{TV}}(\mathbf{x}^{(t)})$
    \State $\mathcal{L}^{(t)} \leftarrow \mathcal{L}_{\text{fidelity}} + \lambda \mathcal{L}_{\text{reg}}$
    \State
    \State \textit{// Backward Pass (Gradient Calculation)}
    \State $\nabla_{\mathbf{y}} \leftarrow (\mathbf{y}^{(t)} - \mathbf{b})$
    \State $\nabla_{\mathbf{x}} \leftarrow \mathcal{A}^T(\nabla_{\mathbf{y}}) + \lambda \nabla_{\mathbf{x}}\mathcal{L}_{\text{reg}}$ \Comment{Algorithm \ref{alg:forward_adjoint}: AdjointOp}
    \State $\nabla_{\mathbf{z}} \leftarrow \nabla_{\mathbf{x}} \odot 2(\mathbf{z}^{(t)} + \epsilon)$ \Comment{Chain rule for NPC}
    \State
    \State \textit{// Parameter Update}
    \State $\eta_t \leftarrow \text{CAWR}(t, \eta_{\min}, \eta_{\max})$ \Comment{Update learning rate}
    \State $\mathbf{z}^{(t+1)} \leftarrow \text{Adam}(\mathbf{z}^{(t)}, \nabla_{\mathbf{z}}, \eta_t)$
\EndFor
\State
\State \textbf{3. Final Output}
\State \Return $\mathbf{x}^* = (\mathbf{z}^{(I_{\max})} + \epsilon)^2$
\end{algorithmic}
\end{algorithm}

\newpage

\section*{Data availability}

The data supporting the findings of this study are provided within the Article. The raw and analysed datasets generated during the study are available for research purposes from the corresponding authors on reasonable request.

\section*{Code availability}

The source code for GPAIR is available at \href{https://github.com/ddffwyb/GPAIR}{https://github.com/ddffwyb/GPAIR}.

\bibliographystyle{unsrt}
\bibliography{references}

\section*{Acknowledgements}

Changhui Li discloses support for the research of this work from the National Key R\&D Program of China (2023YFC2411700 [C.L.], 2017YFE0104200 [C.L.]), the Beijing Natural Science Foundation (7232177 [C.L.]); Chulhong Kim discloses support for the research of this work from the Basic Science Research Program through the National Research Foundation of Korea (NRF) funded by the Ministry of Education (2020R1A6A1A03047902 [C.K.]).

This work is supported by Biomedical Computing Platform of National Biomedical Imaging Center, Peking University.

\section*{Author contributions}

Y.W. conceived and designed the study. S.L. contributed to the methods research and in vivo experiments. T.H. contributed to the results organization and figures preparation. Y.Z. contributed to simulation experiments. C.K. and S.C. provided experimental data. C.L. supervised the study. All authors contributed to the writing of the paper.

\section*{Competing interests}

All authors declare no competing interests.

\end{document}